\documentclass[conference]{IEEEtran}
\IEEEoverridecommandlockouts
\usepackage{cite}
\usepackage{amsmath,amssymb,amsfonts}
\usepackage{algorithmic}
\usepackage{graphicx}
\usepackage{adjustbox}
\usepackage{textcomp}

\usepackage{microtype}
\usepackage{float}

\usepackage{graphicx}
\usepackage{booktabs} 
\usepackage{siunitx}
\usepackage{caption}
\usepackage{url}
\usepackage{hyperref}
\usepackage{multirow, multicol}
\usepackage{rotating}
\usepackage{todonotes}

\usepackage{subcaption}
\usepackage{multirow}
\usepackage{xcolor}

\usepackage{amsmath}
\usepackage{amsthm}
\usepackage{amssymb}

\usepackage{overpic}
\usepackage{bm}

\usepackage{fancyhdr}
\renewcommand{\headrulewidth}{0pt}
\makeatletter 
\newcommand{\linebreakand}{%
  \end{@IEEEauthorhalign}
  \hfill\mbox{}\par
  \mbox{}\hfill\begin{@IEEEauthorhalign}
}
\makeatother 

\usepackage{xcolor}
\def\BibTeX{{\rm B\kern-.05em{\sc i\kern-.025em b}\kern-.08em
    T\kern-.1667em\lower.7ex\hbox{E}\kern-.125emX}}
\begin{document}

\title{ExoTST: Exogenous-Aware Temporal Sequence Transformer for Time Series Prediction}

\author{
\IEEEauthorblockN{Kshitij Tayal\IEEEauthorrefmark{1}\IEEEauthorrefmark{4},
Arvind Renganathan\IEEEauthorrefmark{2}\IEEEauthorrefmark{4},
Xiaowei Jia\IEEEauthorrefmark{3},
Vipin Kumar\IEEEauthorrefmark{2},
and Dan Lu\IEEEauthorrefmark{1}}
\IEEEauthorblockA{\IEEEauthorrefmark{1}Oak Ridge National Laboratory, Oak Ridge, TN, USA}
\IEEEauthorblockA{\IEEEauthorrefmark{2}University of Minnesota, Minneapolis, MN, USA}
\IEEEauthorblockA{\IEEEauthorrefmark{3}University of Pittsburgh, Pittsburgh, PA, USA}
\IEEEauthorblockA{\IEEEauthorrefmark{4}These authors contributed equally}
\IEEEauthorblockA{tayal@umn.edu, renga016@umn.edu, xiaowei@pitt.edu, kumar001@umn.edu, lud1@ornl.gov}
}

\maketitle
\begin{abstract}
Accurate long-term predictions are the foundations for many machine learning applications and decision-making processes.  Traditional time series approaches for prediction often focus on either autoregressive modeling, which relies solely on past observations of the target ``endogenous variables'', or forward modeling, which considers only current covariate drivers ``exogenous variables''. However, effectively integrating past endogenous and past exogenous with current exogenous variables remains a significant challenge. In this paper, we propose ExoTST, a novel transformer-based framework that effectively incorporates current exogenous variables alongside past context for improved time series prediction. To integrate exogenous information efficiently, ExoTST leverages the strengths of attention mechanisms and introduces a novel cross-temporal modality fusion module. This module enables the model to jointly learn from both past and current exogenous series, treating them as distinct modalities. By considering these series separately, ExoTST provides robustness and flexibility in handling data uncertainties that arise from the inherent distribution shift between historical and current exogenous variables. Extensive experiments on real-world carbon flux datasets and time series benchmarks demonstrate ExoTST's superior performance compared to state-of-the-art baselines, with improvements of up to 10\% in prediction accuracy. Moreover, ExoTST exhibits strong robustness against missing values and noise in exogenous drivers, maintaining consistent performance in real-world situations where these imperfections are common.

\end{abstract}

\section{Introduction}



\begin{figure}[t]
        
    \includegraphics[width=1.05\linewidth]{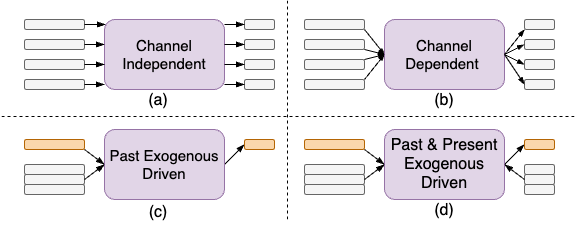}
    
    \vspace{-4pt}
   \caption{\footnotesize We illustrate four classes of problems in time series analysis, differentiated by their handling of exogenous drivers (white) and endogenous response (orange).}
    \label{fig:Compare-Forecasting}
   
\end{figure}

Accurate long-term prediction and forecasting are crucial across various fields, including climate sciences, finance, and biomedicine.
In time series analysis, two prominent modeling approaches are the forward model ($X\to\hat{Y}$) and the autoregressive model ($Y\to\hat{Y}$). The forward model predicts the response ($\hat{Y}$) based on its observed drivers ($X$), making it suitable for understanding the relationship between driver and response. In contrast, the autoregressive model predicts the future values of a time series ($\hat{Y}$) based on its past values ($Y$), which is useful for capturing trends and patterns in the past to make predictions for the future. Figure \ref{fig:Compare-Forecasting} outlines four problems commonly encountered in time series analysis. The first three of these are naturally handled by autoregressive models.  In the first class of problem (Figure 1a), variates do not interact with each other and are treated as being independent. PatchTST \cite{nie2022time}  is an example of an autoregressive method that processes each channel separately without considering dependencies between them. The second class of problems (Figure 1b) deals with scenarios where dependencies exist between input variables. Autoregressive methods like iTransformer \cite{liu2023itransformer} and Crossformer \cite{zhang2022crossformer} are designed to model these dependencies across multiple varieties and incorporate a holistic view of the dynamics within the time series. However, in many multivariate problems, some variables are considered exogenous drivers that affect the endogenous response (target variable). The third class (Figure 1c) of problems is a subset of the second class, where past exogenous and endogenous variates are used to forecast future endogenous values. Methods such as Vector Autoregressive (VAR) \cite{sims1980macroeconomics} models and TimeXer \cite{wang2024timexer} are developed to model these interactions.

For the fourth class of problems in time series analysis (Figure 1d), both past and current/projected exogenous variables are available. However, existing autoregressive approaches for problems modeled in (Figures 1a, 1b, and 1c) fail to leverage all available information, as none of them utilize future exogenous variables for prediction. This paper focuses on methods that are able to leverage past exogenous and endogenous variates alongside current exogenous variates to predict endogenous response. Including current exogenous data for prediction is essential because it captures the immediate influences and conditions that could impact the values of endogenous variates. This approach is particularly useful in numerous scientific applications, where current or future projections of drivers are usually available. For instance, it enables precise prediction of Gross Primary Production (GPP)---the measure of carbon dioxide absorbed by vegetation through photosynthesis---under various climate projections or scenarios (future weather drivers generated by Earth system models). 



Ideally, the methods created to solve the fourth class of problems should retain the autoregressive capability of leveraging historical context while also integrating observed exogenous drivers to predict endogenous responses. This integrated modeling approach can utilize a fuller spectrum of information—both past context and current or projected exogenous drivers—thereby enhancing the robustness and accuracy of predictions. TiDE~\cite{das2023long} is one recent encoder-decoder MLP approach designed to handle the problem illustrated in Figure 1(d). However, such an MLP-based model has deficiencies since it is ill-suited for modeling complex dependencies among the time-series sequences and the covariates, as empirically demonstrated by attention-based papers like iTransformer, TimeXer\cite{liu2023itransformer,wang2024timexer}. 

The alternative methodology used in many scientific applications, such as Earth sciences, for this type of problem, is to build forward models that predict endogenous variables solely on exogenous variables. E.g., LSTM-based forward models are extensively used in hydrology, ecology and climate science\cite{kratzert2018rainfall,willard2022integrating,nathaniel2023metaflux}.  While these forward models utilize exogenous information, they are unable to utilize the historical context effectively.  For instance, when modeling the streamflow response of a hydrological catchment, relying solely on current precipitation (the driver) is insufficient.  It's also essential to integrate past context, which encompasses information about previous precipitation and streamflow patterns, which is crucial for capturing the system's state.

Recognizing the limitations of previous autoregressive approaches (1a, 1b, and 1c) and traditional forward models, this paper presents a novel approach, ExoTST( Exogenous-Aware Temporal Sequence Transformer), that combines information from current or projected exogenous drivers into existing autoregressive approaches. The key challenge driving the development of ExoTST is that, unlike traditional attention-based autoregressive models, our methodology needs to learn not only from past values of the response variable but also to understand the dynamic interactions between drivers and responses. This requires creating models adept at handling long-range, complex dependencies and interactions across time, efficiently capturing the influence of past events and the implications of current external drivers. In this work, we overcome this challenge by drawing inspiration from multimodal models that learn joint representations of diverse and noisy data such as text, images, and videos. These models utilize a combination of cross-attention and self-attention mechanisms to project insights from one modality into another, facilitating a better understanding across diverse data types. Building on these concepts, our approach, ExoTST, integrates past context as distinct modalities while treating the endogenous response as another modality. Within our framework, we implement patch-wise self-attention to capture temporal relationships in both endogenous and exogenous variables effectively. ExoTST comprises of two encoders dedicated to processing past and current exogenous modalities and a decoder for the endogenous target series. A cross-temporal fusion module facilitates efficient information exchange between exogenous encodings through cross-attention over aggregate tokens. Subsequently, the endogenous decoder incorporates the fused exogenous information and past endogenous values via cross-attention to generate predictions. By explicitly modeling these drivers and integrating diverse data modalities, our proposed architecture not only maintains the historical context but also adapts to new inputs.   Our contributions are summarized below:
\begin{itemize}
    \item In this paper, we highlight the importance of incorporating current/projected exogenous information and present ExoTST that improves upon the existing autoregressive approaches for cases where current/projected exogenous drivers are available. By incorporating these exogenous drivers, ExoTST systematically captures and integrates external influences into the forecasting model, significantly improving prediction accuracy.  
    \item We conduct extensive experiments on real-world climate and benchmark datasets and show that ExoTST outperforms existing forward and autoregressive methods by 8-12\%. Furthermore, we demonstrate that ExoTST maintains robust predictive accuracy even in realistic scenarios where future exogenous variables are derived from potentially noisy forecasts.
\end{itemize}

\section{Related Work}

\textit{Autoregressive Time Series Approaches}: 
Autoregressive model predicts the future values of a response variable based on its past values. Recently, Transformers-based models such as Autoformer \cite{wu2021autoformer}, PatchTST \cite{nie2022time}, Crossformer \cite{zhang2022crossformer} have garnered significant interest in the autoregressive forecast due to their ability to capture long-term temporal dependencies and complex multivariate correlations. These models have been adapted and modified to address the unique challenges posed by time series, such as capturing long-term dependencies, handling multivariate data, and efficiently processing large-scale datasets. For eg.  Autoformer \cite{wu2021autoformer}, uses the concept of Auto-correlation to discover sub-series similarity as a replacement for the canonical self-attention mechanism. However, unlike the point-wise time series approaches in Autoformer, PatchTST \cite{nie2022time} takes a different approach by splitting these time series data into subseries-level patches and then capturing dependencies between these patches. This helps them in extracting more meaningful local patterns and relationships by making them more computationally efficient. Crossformer \cite{zhang2022crossformer} takes a similar approach of patching while additionally using a two-stage attention layer to efficiently capture the cross-variate dependencies on top of dependencies across time for each patch. iTransformer \cite{liu2023itransformer} captures an expanded receptive field by increasing the patch length from smaller segments to the entire variate, producing a global representation of each variate. The model then applies variate-wise cross attention to capture multivariate correlations. Recently, another family of MLP-based approaches like Dlinear \cite{zeng2023transformers},  NHITS \cite{challu2023nhits}, 	FreTS \cite{yi2024frequency} have become popular for being more efficient while remaining competitive with transformer models for autoregressive forecasting. However, most of these autoregressive approaches primarily focus on univariate or multivariate forecasting predominantly based on past data, often ignoring the effect of exogenous drivers on the current endogenous response of the system. This is a key distinction from the setting described in this paper. 

\textit{Time series prediction with exogenous variables}: In many multivariate scenarios, certain variables, known as exogenous drivers, play a crucial role in influencing the prediction of the endogenous response or target variable. Numerous methodologies have been proposed for time series prediction and forecasting that aim to leverage the historical context comprising past exogenous drivers and endogenous responses. Notable methods include Vector Autoregressive (VAR) models \cite{sims1980macroeconomics} and the TimeXer approach \cite{wang2024timexer}, which have been developed specifically to model these interactions. Furthermore, there's been an exploration into time series prediction techniques incorporating current or projected exogenous variables, leveraging both historical context and present or projected exogenous variables. These methods leverage a system's past context—including past drivers(exogenous)  and responses (endogenous). —and its current or projected exogenous variables. One such approach is the Temporal Fusion Transformer (TFT) \cite{lim2021temporal}, which integrates the strengths of Transformers, RNNs, and Gating Mechanisms. TFT utilizes both the past context and current exogenous variables(drivers).  Similarly, NBEATSx \cite{olivares2023neural} extends the autoregressive N-BEATS architecture to accommodate exogenous drivers. By incorporating a dedicated branch for processing these variables (drivers), NBEATSx improves prediction accuracy and interoperability. In recent years, TiDE \cite{das2023long}, an MLP-based approach, has emerged as a state-of-the-art model for leveraging current or projected exogenous variables on contemporary benchmarks. However, a challenge arises with TiDE as it concatenates exogenous drivers with endogenous features at each time point, necessitating alignment of the endogenous and exogenous variables in time. Real-world time series often suffer from issues like missing values and uneven sampling, which pose significant challenges in modeling the effects of exogenous variables on endogenous variables using TiDE.  Additionally, MLP-based approaches may exhibit limitations in capturing complex dependencies among time-series sequences and covariates\cite{liu2023itransformer,wang2024timexer}. In contrast, our approach, ExoTST, introduces external information through innovative embedding, cross-attention, and cross-temporal fusion techniques, all within an attention-based framework. This enables the effective incorporation of external information into patch-wise representations of endogenous variables, facilitating adaptation to time-lagged or data-missing records.

\section{Problem Formulation and Data Description}

\subsection{Problem Description}

Given an endogenous univariate time series $\mathbf{y}_{1:L} = \{y_1, y_2, \dots, y_L\}$, and corresponding exogenous drivers $\mathbf{X}_{1:L} = \{\mathbf{x}_1, \mathbf{x}_2, \dots, \mathbf{x}_L\}$ over the past $L$ time steps, 
we aim to predict future values $\hat{y}_{L+1:L+f} = \{\hat{y}_{L+1}, \hat{y}_{L+2}, \dots, \hat{y}_{L+f}\}$. This prediction task is based on the current/projected exogenous drivers 
$\mathbf{X}_{L+1:L+f} = \{\mathbf{x}_{L+1}, \mathbf{x}_{L+2}, \dots, \mathbf{x}_{L+f}\}$. 
Here the indices $1$ to $L$ represent the total number of past observations (past context), and $L+1$ to $L+f$ indicate the exogenous data available during forecasting, where $f$ represents the length of the prediction window. Each $\mathbf{x}_t$ is a multivariate observation at time $t$, consisting of $M$ distinct exogenous drivers, $\mathbf{x}_t = (x_t^1, x_t^2, \dots, x_t^M)$. Notably, each $\mathbf{x}_t$ may contain missing values for certain exogenous drivers. 
In other words, the number of available observations for drivers can vary, accounting for missing data. The objective is to construct a model $f$, that can efficiently handle problem in Figure 1d, which can be formalized as: $\hat{y}_{L+1:L+f} = f(\mathbf{y}_{1:L}, \mathbf{X}_{1:L}, \mathbf{X}_{L+1:L+f})$

\section{Our approach: ExoTST}

\begin{figure*}[!t]
	\centering
	\begin{overpic}[width=0.80\textwidth]{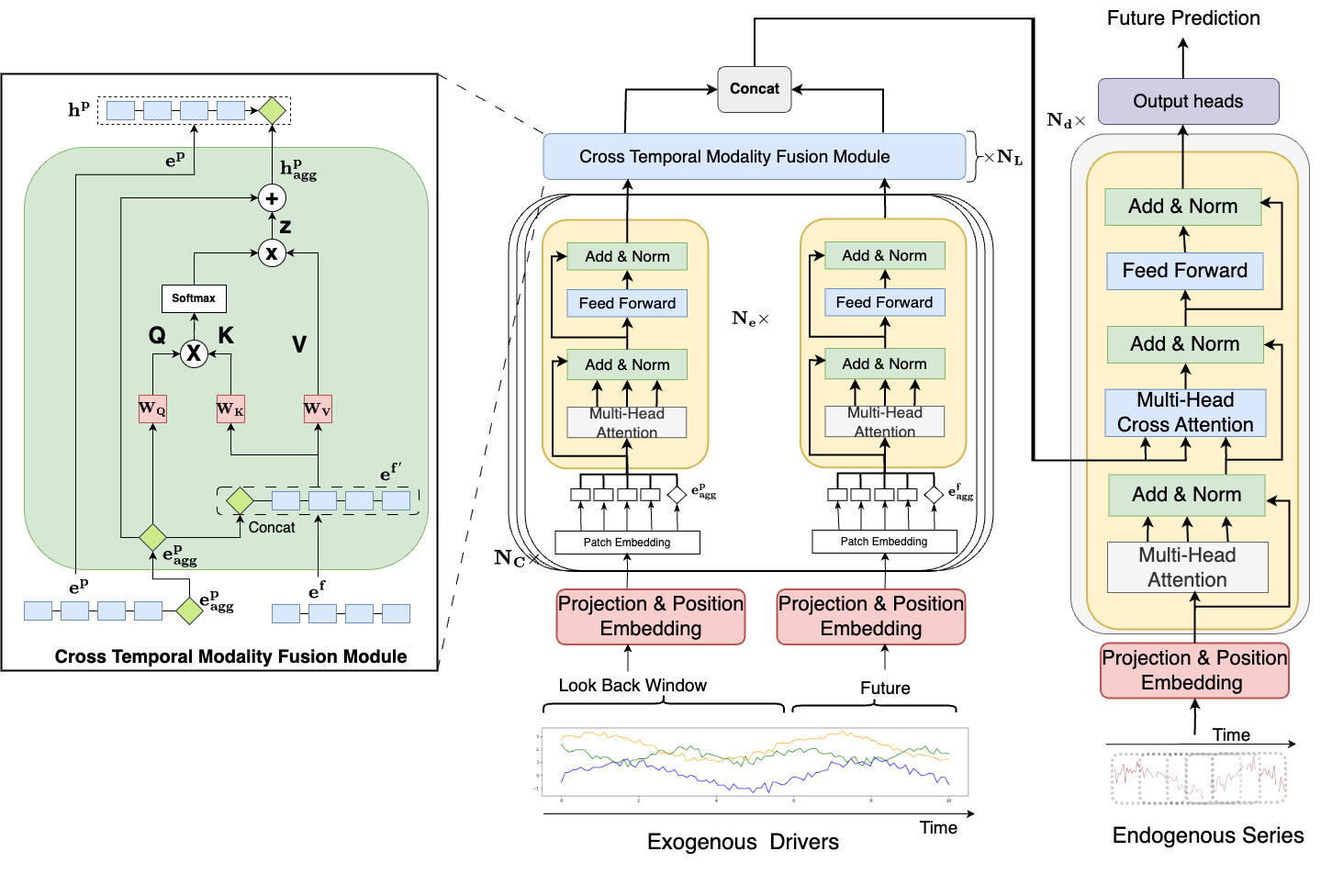} 
	\end{overpic}
	\caption{\small ExoTST Architecture for time series prediction, featuring cross-temporal modality fusion and multi-head attention to integrate and predict based on past and future exogenous drivers. }\label{fig:future_tst_architecture}
\end{figure*}

Our proposed method (see Figure \ref{fig:future_tst_architecture}) builds upon transformer-based architecture due to its effectiveness in handling sequences, managing complex data dependencies, and scaling to large time series data. Our framework consists of an encoder-decoder approach as utilized in multimodal LLMs. Here, the encoder and decoder need to extract local semantic information from an input series, which is essential for analyzing dependencies within the time series. To extract local semantic information, we employ patch-wise encoding for all input series (both endogenous and exogenous).  We use two encoders to process the past exogenous and current/projected exogenous series. We then propose a cross-temporal fusion module \cite{chen2021crossvit} to align and denoise these modalities (past and current/projected) through information exchange. The resultant exogenous encoding is then fed as an input to the decoder. In the following, we present an overview of the \textit{patching mechanism}, detailing its ability to reduce time and space complexity. We then discuss our \textit{exogenous encoders}, designed to manage past and future series, alongside the \textit{cross-temporal modality fusion module}, which enables efficient information exchange between these series. Finally, we will describe how the \textit{endogenous decoder} utilizes all the gathered information to forecast future time series.

\subsection{Patching Time Series}
Patching helps in grouping local time steps that may contain similar values. We adopt a representation that divides the input series $\boldsymbol{x}^{(i)}$ into overlapping segments \cite{nie2022time}, where $(i)$ represents the index of a specific variate. Each segment of the series is then mapped to a unique temporal token (patch)
through a trainable patch embedding module. We denote the series length of the input as $I$ (which will be different for encoder/decoder), patch length as $P$, and the stride - the nonoverlapping region between two consecutive patches - as $S$. The patching process generates the sequence of patches $\boldsymbol{x}_p^{(i)} \in \mathbb{R}^{P \times N}$ as follows:
\begin{equation}
\boldsymbol{x}^{(i)} \rightarrow \boldsymbol{x}_p^{(i)} \in \mathbb{R}^{P \times N}, \quad \text{where } N = \left\lfloor\frac{(I-P)}{S}\right\rfloor + 2.
\end{equation}

Here, we pad $S$ 
by 
repeatedly adding the last value
$x_I^{(i)} \in \mathbb{R}$ to the end of the original sequence to complete the last patch of length $P$. 
To address the shift in data distribution between training and testing, we also implement instance normalization by normalizing each time series to have a zero mean and unit standard deviation before patching and adding them later. These patches are then mapped to the transformer latent space of dimension $D$ via a trainable linear projection $\mathbf{W}_p \in \mathbb{R}^{D \times P}$. To incorporate the sequential nature of the input data, we also incorporate additive learnable sinusoidal position encoding $\mathbf{W}_{\text {pos}} \in \mathbb{R}^{D \times N}$, which enable our model to consider the order of tokens in the absence of recurrent structures. This is modeled as:
$\boldsymbol{x}_d^{(i)} = \mathbf{W}_{p} \cdot \boldsymbol{x}_{\text{p}}^{(i)} + \mathbf{W}_{\text{pos}}$.


In our model, we also introduce an aggregation token, termed $e_{\text{agg}}$, to both the past and future patch embeddings of exogenous drivers, similar to the approach used in BERT \cite{devlin2018bert} and ViT \cite{dosovitskiy2020image}. This $e_{\text{agg}}$ token interacts with the patch tokens at every encoder layer, serving as an agent that encapsulates the information from all patch tokens in a modality. This mechanism proves especially beneficial in the  Cross Temporal Fusion Module, which will be detailed later. 
We conduct patching in similar ways for both exogenous and endogenous time series. 
Utilizing patches significantly reduces the number of input tokens from $L$ to approximately $L / S$. which simplifies the attention maps' computational complexity by a factor of $S$. As a result, ExoTST can learn from a broader lookback window while avoiding additional memory and computational overhead.

\subsection{Exogenous Encoders}
Capturing the temporal relationships within exogenous data series is crucial for accurate prediction and understanding of the underlying dynamics. To address this challenge, we propose the use of two attention-based encoders \cite{vaswani2017attention}, one for past exogenous series and another for current/projected exogenous series. By treating the past and projected exogenous series as different modalities, our approach offers flexibility and robustness in handling noise and uncertainties in exogenous data. 
The input to each encoder is a matrix $\boldsymbol{x}_d^{(i)} \in \mathbb{R}^{D \times (N+1)}$, where $D$ represents the dimension of latent embedding, and $N$ denotes the number of patches, and the additional 1 accounts for the $e_{\text{agg}}$ token. The number of patches differs between the two encoders, depending on the look-back window and forecast horizon. By first applying self-attention to the input patches, the encoders learn the temporal dependencies and capture the overall dynamics and interactions between different patches.

After the initial embedding from patching, the model employs a multi-head self-attention mechanism to attend to different parts of the sequence simultaneously, capturing various aspects of the input at different positions.  For each head $h = 1, \ldots, H$, the transformations are: 
\begin{equation}
\begin{aligned}
    Q_h^{(i)}, K_h^{(i)}, V_h^{(i)} &= (\boldsymbol{x}_d^{(i)})^T \mathbf{W}_h^Q, \;(\boldsymbol{x}_d^{(i)})^T \mathbf{W}_h^K, \;(\boldsymbol{x}_d^{(i)})^T \mathbf{W}_h^V, \\
    \end{aligned}
\label{eq:attention}
\end{equation}
where $d_k = \dfrac{D}{H}$,
 \(\mathbf{W}_h^Q, \mathbf{W}_h^K \in \mathbb{R}^{D \times d_k}\), and \(\mathbf{W}_h^V \in \mathbb{R}^{D \times D}\). Additionally, \(Q_h^{(i)}, K_h^{(i)} \in \mathbb{R}^{N \times d_k}\), and \(V_h^{(i)} \in \mathbb{R}^{N \times D}\). Afterwards, attention mechanism within each head is computed as follows:
\begin{equation}
\operatorname{Attention}\left(Q_h^{(i)}, K_h^{(i)}, V_h^{(i)}\right)=\operatorname{Softmax}\left(\frac{Q_h^{(i)} K_h^{(i)^T}}{\sqrt{d_k}}\right) V_h^{(i)}
\label{eq:self_attention}
\end{equation}

The multi-head attention block enhances our model's ability to focus on different positions in various contexts. This block also includes BatchNorm layers and a position-wise feed-forward network that applies two linear transformations with a ReLU activation and residual connections, as shown in Figure \ref{fig:future_tst_architecture}. The feed-forward component allows the encoder to learn non-linear transformations, increasing its expressive power.
To stabilize the learning process and improve training efficiency, each sub-layer (both attention and feed-forward networks) also includes a residual connection followed by layer normalization. We employ $N_e$ such identical layers to process input sequences, efficiently transforming them into higher-level representations. This facilitates parallel processing and scales effectively with increased data and model size, addressing the limitations of previous sequence prediction models like RNNs and LSTMs.

\subsection{Cross Temporal Modality Fusion Module}
ExoTST processes past and current/projected exogenous series tokens by two separate encoders. To integrate and denoise these tokens, we propose a cross temporal modeling fusion module, which fuses tokens from both encoders to complement each other. To enhance efficiency, we utilize token fusion methodology based on cross-attention \cite{chen2021crossvit,yan2022multiview}, which uses a single token from each encoder as a query to exchange information with the other encoder. Figure \ref{fig:future_tst_architecture} illustrates the network architecture of our proposed Cross-Temporal Modality Fusion Module. The proposed module combines the temporal embedding from the look-back encoder with other encoders, and vice versa; both processes are identical. Here, we detail the base integration process with the help of the look back encoder. Specifically, we take aggregate embedding from the lookback encoder ($e^p_{\text{agg}}$) and concatenate it with the patch embeddings from the other encoder ($e^f$), forming a new embedding $e^{f'}$:
\begin{equation}
e^{f'} = \left[ e^{p}_{\text{agg}} \parallel e^f \right]
\end{equation}
The fusion module then performs a cross-attention operation between the aggregate token $e^p_{\text{agg}}$ and the concatenated embedding $e^{f'}$. In this operation, $e^p_{\text{agg}}$ serves as the query, while $e^{f'}$ provides the keys and values. The following transformations are applied, where $\mathbf{W}_Q$, $\mathbf{W}_K$, and $\mathbf{W}_V$ are the weight matrices for the query, key, and value, respectively:
\begin{align}
\left[\mathbf{Q}, \mathbf{K}, \mathbf{V}\right] &= \left[e^p_{\text{agg}} \mathbf{W}_Q,; e^{f'} \mathbf{W}_K, ;e^{f'} \mathbf{W}_V\right] \\
\operatorname{CA}(\mathbf{z}) &= \operatorname{softmax}\left(\frac{\mathbf{QK}^\top}{\sqrt{D_h}}\right) \mathbf{V}\
\end{align}
By utilizing $e^p_{\text{agg}}$ exclusively for queries, both computational and memory complexities of the attention map $\mathbf{A}$ are reduced to linear, as opposed to the quadratic complexity seen in full attention mechanisms. Furthermore, the cross-attention employs multiple heads (MCA) to enhance its ability to capture diverse temporal features without requiring a subsequent feed-forward network. The final output, incorporating layer normalization and residual connections, is computed as follows:
\begin{align}
h^p_{\text{agg}} &= e^p_{\text{agg}} + \text{MCA}\left(\text{LN}\left(e^{f'}\right) \right) \\
h^{p} &= \left[ h^p_{\text{agg}} \parallel e^p \right]
\end{align}
The aggregate embedding from the future encoder ($e^f_{\text{agg}}$) undergoes a similar process, with the roles of the lookback and future embeddings reversed to produce $h^p_{\text{agg}}$. The fusion module is applied $N_L$ times to facilitate the exchange of information between the two encoders at different levels of abstraction. This iterative fusion process allows the model to learn rich, complementary features from both the lookback and future time frames.
By leveraging the aggregation token $e_{\text{agg}}$ at each encoder output as an agent for information exchange, our proposed cross-temporal modality fusion module efficiently integrates temporal features from different time scales. The $e_{\text{agg}}$ token, which captures abstract information among all patch tokens within its branch, interacts with patch tokens from the other branch to incorporate information. Subsequently, the updated $e_{\text{agg}}$ token passes the learned information from the other branch to its own patch tokens in the next layer, enriching the representation of each patch token.
The proposed cross-temporal modality fusion module offers an efficient and effective approach to integrating temporal information from both past and future time frames. By enabling the exchange of information at different levels of abstraction, this module enhances the model's ability to learn robust feature representations suitable for time series prediction tasks. The outputs from the fusion module from both modalities are then concatenated together and reshaped to produce the final output \textit{O'}, which is fed to the decoder module.

\begin{align}
\textit{O} &= \left[ h^p_{\text{agg}} \parallel h^f_{\text{agg}} \right] \in \mathbb{R}^{C \times ((N' + 1) + (N'' + 1)) \times D} \\
\textit{O'} &= \text{Reshape}(\textit{O'}) \in \mathbb{R}^{1 \times C((N' + 1) + (N'' + 1)) \times D}
\end{align}

\subsection{Endogenous Decoder} 

Our decoder architecture consists of $N_d$ identical layers, each comprising three primary sub-layers: multi-head self-attention, multi-head cross-attention, and position-wise feed-forward networks. The decoder processes endogenous series, taking $\boldsymbol{y}_p^{(i)} \in \mathbb{R}^{P \times N}$ as input, where $P$ represents the number of time steps and $N$ denotes the number of patches. The decoder generates a representation denoted as $\boldsymbol{z}^{(i)} \in \mathbb{R}^{D \times N}$, where $D$ is the decoder latent dimension, which is the same as the encoder latent dimension. We don't use aggregation token ($e_{\text{agg}}$) with the decoder.

The first step in the decoder involves applying multi-head self-attention, as described in equations \ref{eq:attention} and \ref{eq:self_attention}, which enables the decoder to model the temporal relationships within the endogenous series effectively. Following the self-attention layer, the decoder employs a multi-head cross-attention layer, which is crucial for learning the relationships between exogenous drivers and endogenous responses. The endogenous output from the self-attention layer serves as queries, while the output of the cross-temporal modality fusion module embedding acts as the key and value. This cross-attention mechanism allows the decoder to integrate contextual exogenous information when generating endogenous predictions. By propagating the multivariate correlations learned by the exogenous embedding to the endogenous temporal tokens, the cross-attention layer facilitates information transfer between the two types of modalities, a common approach in multi-modal learning.
In addition to the attention layers, the decoder block includes BatchNorm layers and a position-wise feed-forward network. The feed-forward network applies two linear transformations with a ReLU activation and residual connections, similar to the encoder, to stabilize training and learn non-linear transformations. Finally, a linear projection is applied to the decoder output to obtain the prediction result $\hat{\boldsymbol{y}}^{(i)}=$ $\left(\hat{y}_{L+1}^{(i)}, \ldots, \hat{y}_{L+F}^{(i)}\right) \in \mathbb{R}^{1 \times F}$. To measure the discrepancy between the predicted and ground truth values, we employ the Mean Squared Error (MSE) loss. The loss is gathered and averaged over all samples to obtain the overall objective loss: $\mathcal{L}=\mathbb{E}_{\boldsymbol{y}} \left\|\hat{\boldsymbol{y}}_{L+1: L+F}^{(i)}-\boldsymbol{y}_{L+1: L+F}^{(i)}\right\|_2^2$.

By leveraging these mechanisms, including self-attention, cross-attention, and feed-forward networks, the endogenous decoder maintains sequential fidelity and context sensitivity. These properties are essential for our prediction task, as they enable the decoder to effectively capture the temporal dependencies within the endogenous series and integrate relevant exogenous information to generate accurate predictions. The attention decoder's ability to model complex relationships and maintain context awareness makes it a powerful component in our multivariate time series prediction architecture.

   
\section{Results}
To evaluate our proposed approach for long-term temporal modeling, we perform comprehensive experiments on three real-world eddy-covariance flux tower sites: DE-HAI (deciduous forest, Germany), ES-LJU (open shrubland, Spain), and AT-NEU (managed grassland, Austria), as well as three benchmark datasets: electricity, ETTh1, and Exchange. 

\begin{figure}[t]
    \centering
    \includegraphics[width=0.9\linewidth]{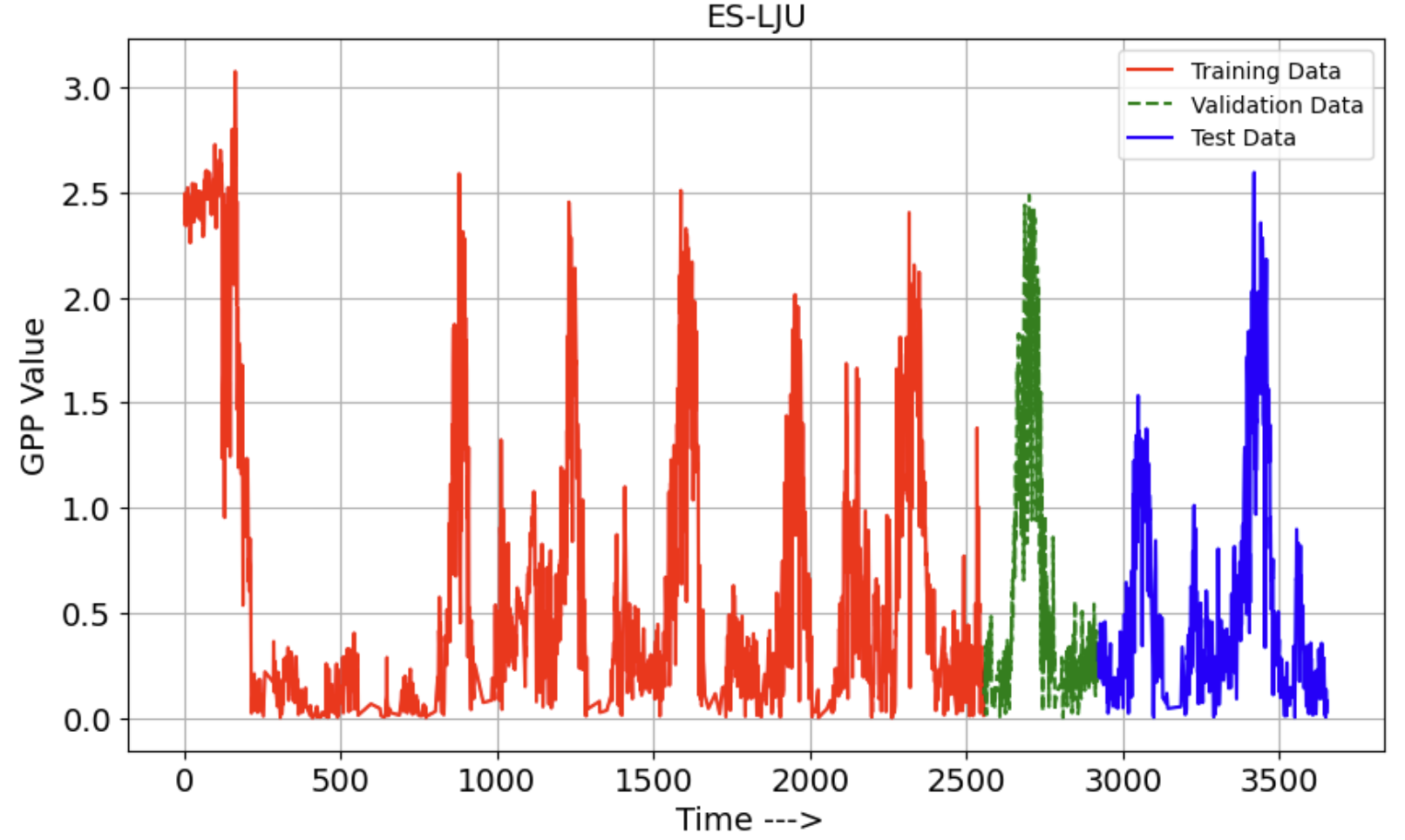} 
    
    \caption{\footnotesize This plot visualizes daily GPP (Gross Primary Productivity) for ES-LJU with training, validation, and test data separated by color (red, green, blue).}
    \label{fig:gpp_illustration}
    
\end{figure}
\textbf{Baseline}: Our baseline comprises state-of-the-art autoregressive transformer-based models ${Y\to\hat{Y}}$, including univariate PatchTST \cite{nie2022time} (focused solely on past endogenous series for future predictions), (M)PatchTST \cite{nie2022time} (utilizing past exogenous and endogenous series to forecast both future endogenous and exogenous series), and iTrans \cite{liu2023itransformer} (leveraging past exogenous and endogenous series for forecasting both future endogenous and exogenous series). Additionally, we include DLinear as an important baseline due to its simplicity and effectiveness. We further compare with two LSTM-based forward models ${X\to\hat{Y}}$: (a) Encoder-Decoder (ED-LSTM)\cite{yin2022rr} incorporates past context, (b) LSTM\cite{kratzert2018rainfall} without past context.  We also include MLP-based TiDE \cite{das2023long} model. TiDE and the ED-LSTM are the only baselines capable of utilizing current or projected exogenous inputs alongside past context, whereas LSTM solely relies on current or projected exogenous inputs. In total, we compared our method against seven time series baselines.
All models are trained using Mean Squared Error (MSE) as the training loss. The train:validation:test ratio across all datasets follows the convention of 7:1:2, as recommended by prior work \cite{wu2021autoformer}. Note that all experiments are conducted on standard normalized datasets, employing the mean and standard deviations from the training period. This approach ensures consistency with previous methodologies \cite{das2023long}.

\textbf{Our Model} We use the architecture describe in Figure \ref{fig:future_tst_architecture}. Our model features a fixed lookback window \(L = 256\), attention dimension \(D = 256\), patch length \(P = 16\), and stride \(S = 8\). It incorporates an 8-head multihead attention mechanism in both encoder and decoder stages. The MLP size is set to 128, and the architecture includes two layers for both encoder and decoder. We adjusted our hyperparameters using the validation set, implementing early stopping when the loss did not decrease for 10 consecutive epochs. The model optimization was conducted using the Adam Optimizer with a learning rate of \(1 \times 10^{-4}\). We utilized the default configurations for all baseline models.

\begin{table*}[htbp]
\caption{Full results of the long-term prediction task for carbon flux with exogenous drivers. The lowest MSE for each horizon and dataset is \textbf{emphasized} in bold.}
\vspace{2pt}
\setlength{\tabcolsep}{2.5pt}
\renewcommand{\arraystretch}{1.5}
\begin{sc}
\resizebox{\linewidth}{!}{
\begin{tabular}{c|c|cc|cc|cc|cc|cc|cc|cc|cc|cc}
\toprule[1.2pt]  
\multicolumn{2}{c}{\multirow{2}{*}{\scalebox{1.35}{Models}}}   & \multicolumn{2}{c}{\scalebox{1.35}{ExoTST}} & \multicolumn{2}{c}{\scalebox{1.35}{iTrans.}} & \multicolumn{2}{c}{\scalebox{1.35}{(U)PatchTST}} & \multicolumn{2}{c}{\scalebox{1.35}{(M)PatchTST}} & \multicolumn{2}{c}{\scalebox{1.35}{DLinear} }& \multicolumn{2}{c}{\scalebox{1.35}{TiDE}} & \multicolumn{2}{c}{\scalebox{1.35}{ED-LSTM}} & \multicolumn{2}{c}{\scalebox{1.35}{LSTM}} \\ 
\multicolumn{2}{c}{} & \multicolumn{2}{c}{\scalebox{1.35}{(Ours)}} & \multicolumn{2}{c}{\scalebox{1.35}{\cite{liu2023itransformer}}} & \multicolumn{2}{c}{\scalebox{1.35}{\cite{nie2022time}}} & \multicolumn{2}{c}{\scalebox{1.35}{\cite{nie2022time}}} & \multicolumn{2}{c}{\scalebox{1.35}{\cite{zeng2023transformers}}} & \multicolumn{2}{c}{\scalebox{1.35}{\cite{das2023long}}} & \multicolumn{2}{c}{\scalebox{1.35}{\cite{yin2022rr}}} & \multicolumn{2}{c}{\scalebox{1.35}{\cite{kratzert2018rainfall}}}\\ 
\cmidrule(lr){3-4}\cmidrule(lr){5-6}\cmidrule(lr){7-8}\cmidrule(lr){9-10}\cmidrule(lr){11-12}\cmidrule(lr){13-14}\cmidrule(lr){15-16}\cmidrule(lr){17-18}
\multicolumn{2}{c}{\scalebox{1.35}{Metric}}   & \scalebox{1.35}{MSE} & \scalebox{1.35}{MAE} & \scalebox{1.35}{MSE}    & \scalebox{1.35}{MAE}   & \scalebox{1.35}{MSE}    & \scalebox{1.35}{MAE}    & \scalebox{1.35}{MSE}    & \scalebox{1.35}{MAE} & \scalebox{1.35}{MSE}    & \scalebox{1.35}{MAE} & \scalebox{1.35}{MSE}    & \scalebox{1.35}{MAE}   & \scalebox{1.35}{MSE}    & \scalebox{1.35}{MAE}  & \scalebox{1.35}{MSE}    & \scalebox{1.35}{MAE}      \\ 
\toprule[1.2pt] 
\multirow{8}{*}{\scalebox{1.35}{\rotatebox{90}{DE-HAI}}}
 & \scalebox{1.35}{1} & \scalebox{1.35}{\textbf{0.067}} & \scalebox{1.35}{0.169} & \scalebox{1.35}{0.076} & \scalebox{1.35}{0.194} & \scalebox{1.35}{0.074} & \scalebox{1.35}{0.186} & \scalebox{1.35}{0.080} & \scalebox{1.35}{0.205} & \scalebox{1.35}{0.071} & \scalebox{1.35}{0.179} & \scalebox{1.35}{\textbf{0.067}} & \scalebox{1.35}{0.161} & \scalebox{1.35}{0.084} & \scalebox{1.35}{0.212} & \scalebox{1.35}{0.519} & \scalebox{1.35}{0.658} \\
 & \scalebox{1.35}{7} & \scalebox{1.35}{\textbf{0.073}} & \scalebox{1.35}{0.189} & \scalebox{1.35}{0.074} & \scalebox{1.35}{0.186} & \scalebox{1.35}{0.076} & \scalebox{1.35}{0.192} & \scalebox{1.35}{0.079} & \scalebox{1.35}{0.199} & \scalebox{1.35}{0.078} & \scalebox{1.35}{0.196} & \scalebox{1.35}{0.078} & \scalebox{1.35}{0.196} & \scalebox{1.35}{0.108} & \scalebox{1.35}{0.254} & \scalebox{1.35}{0.305} & \scalebox{1.35}{0.452} \\
 & \scalebox{1.35}{14} & \scalebox{1.35}{\textbf{0.076}} & \scalebox{1.35}{0.189} & \scalebox{1.35}{0.077} & \scalebox{1.35}{0.189} & \scalebox{1.35}{0.082} & \scalebox{1.35}{0.201} & \scalebox{1.35}{0.080} & \scalebox{1.35}{0.197} & \scalebox{1.35}{0.085} & \scalebox{1.35}{0.209} & \scalebox{1.35}{0.098} & \scalebox{1.35}{0.225} & \scalebox{1.35}{0.106} & \scalebox{1.35}{0.249} & \scalebox{1.35}{0.240} & \scalebox{1.35}{0.390} \\
 & \scalebox{1.35}{30} & \scalebox{1.35}{\textbf{0.077}} & \scalebox{1.35}{0.186} & \scalebox{1.35}{0.083} & \scalebox{1.35}{0.200} & \scalebox{1.35}{0.089} & \scalebox{1.35}{0.214} & \scalebox{1.35}{0.087} & \scalebox{1.35}{0.208} & \scalebox{1.35}{0.094} & \scalebox{1.35}{0.224} & \scalebox{1.35}{0.098} & \scalebox{1.35}{0.228} & \scalebox{1.35}{0.124} & \scalebox{1.35}{0.265} & \scalebox{1.35}{0.199} & \scalebox{1.35}{0.345} \\
 & \scalebox{1.35}{60} & \scalebox{1.35}{\textbf{0.088}} & \scalebox{1.35}{0.210} & \scalebox{1.35}{0.094} & \scalebox{1.35}{0.222} & \scalebox{1.35}{0.095} & \scalebox{1.35}{0.218} & \scalebox{1.35}{0.097} & \scalebox{1.35}{0.224} & \scalebox{1.35}{0.104} & \scalebox{1.35}{0.238} & \scalebox{1.35}{0.106} & \scalebox{1.35}{0.238} & \scalebox{1.35}{0.140} & \scalebox{1.35}{0.286} & \scalebox{1.35}{0.189} & \scalebox{1.35}{0.333} \\
 & \scalebox{1.35}{90} & \scalebox{1.35}{\textbf{0.095}} & \scalebox{1.35}{0.221} & \scalebox{1.35}{0.107} & \scalebox{1.35}{0.243} & \scalebox{1.35}{0.104} & \scalebox{1.35}{0.231} & \scalebox{1.35}{0.106} & \scalebox{1.35}{0.236} & \scalebox{1.35}{0.110} & \scalebox{1.35}{0.250} & \scalebox{1.35}{0.113} & \scalebox{1.35}{0.248} & \scalebox{1.35}{0.155} & \scalebox{1.35}{0.302} & \scalebox{1.35}{0.174} & \scalebox{1.35}{0.321} \\
 & \scalebox{1.35}{120} & \scalebox{1.35}{\textbf{0.115}} & \scalebox{1.35}{0.256} & \scalebox{1.35}{0.124} & \scalebox{1.35}{0.272} & \scalebox{1.35}{0.117} & \scalebox{1.35}{0.252} & \scalebox{1.35}{0.123} & \scalebox{1.35}{0.260} & \scalebox{1.35}{0.131} & \scalebox{1.35}{0.274} & \scalebox{1.35}{0.118} & \scalebox{1.35}{0.255} & \scalebox{1.35}{0.168} & \scalebox{1.35}{0.317} & \scalebox{1.35}{0.181} & \scalebox{1.35}{0.330} \\
\cmidrule(lr){2-18}
 & \scalebox{1.35}{AVG} & \scalebox{1.35}{\textbf{0.084}} & \scalebox{1.35}{0.203} & \scalebox{1.35}{0.091} & \scalebox{1.35}{0.215} & \scalebox{1.35}{0.091} & \scalebox{1.35}{0.213} & \scalebox{1.35}{0.093} & \scalebox{1.35}{0.218} & \scalebox{1.35}{0.096} & \scalebox{1.35}{0.224} & \scalebox{1.35}{0.097} & \scalebox{1.35}{0.222} & \scalebox{1.35}{0.126} & \scalebox{1.35}{0.269} & \scalebox{1.35}{0.258} & \scalebox{1.35}{0.404} \\
\bottomrule[1.2pt]
\multirow{8}{*}{\scalebox{1.35}{\rotatebox{90}{AT-NEU}}}
 & \scalebox{1.35}{1} & \scalebox{1.35}{\textbf{0.162}} & \scalebox{1.35}{0.285} & \scalebox{1.35}{0.238} & \scalebox{1.35}{0.339} & \scalebox{1.35}{0.210} & \scalebox{1.35}{0.331} & \scalebox{1.35}{0.185} & \scalebox{1.35}{0.315} & \scalebox{1.35}{0.166} & \scalebox{1.35}{0.287} & \scalebox{1.35}{0.163} & \scalebox{1.35}{0.286} & \scalebox{1.35}{0.228} & \scalebox{1.35}{0.355} & \scalebox{1.35}{0.764} & \scalebox{1.35}{0.751} \\
 & \scalebox{1.35}{7} & \scalebox{1.35}{\textbf{0.255}} & \scalebox{1.35}{0.348} & \scalebox{1.35}{0.281} & \scalebox{1.35}{0.367} & \scalebox{1.35}{0.271} & \scalebox{1.35}{0.378} & \scalebox{1.35}{0.273} & \scalebox{1.35}{0.377} & \scalebox{1.35}{0.277} & \scalebox{1.35}{0.382} & \scalebox{1.35}{0.278} & \scalebox{1.35}{0.370} & \scalebox{1.35}{0.282} & \scalebox{1.35}{0.396} & \scalebox{1.35}{0.475} & \scalebox{1.35}{0.544} \\
 & \scalebox{1.35}{14} & \scalebox{1.35}{\textbf{0.248}} & \scalebox{1.35}{0.356} & \scalebox{1.35}{0.297} & \scalebox{1.35}{0.377} & \scalebox{1.35}{0.303} & \scalebox{1.35}{0.402} & \scalebox{1.35}{0.292} & \scalebox{1.35}{0.389} & \scalebox{1.35}{0.317} & \scalebox{1.35}{0.415} & \scalebox{1.35}{0.334} & \scalebox{1.35}{0.409} & \scalebox{1.35}{0.306} & \scalebox{1.35}{0.411} & \scalebox{1.35}{0.403} & \scalebox{1.35}{0.498} \\
 & \scalebox{1.35}{30} & \scalebox{1.35}{\textbf{0.289}} & \scalebox{1.35}{0.386} & \scalebox{1.35}{0.319} & \scalebox{1.35}{0.395} & \scalebox{1.35}{0.317} & \scalebox{1.35}{0.407} & \scalebox{1.35}{0.330} & \scalebox{1.35}{0.417} & \scalebox{1.35}{0.353} & \scalebox{1.35}{0.444} & \scalebox{1.35}{0.339} & \scalebox{1.35}{0.431} & \scalebox{1.35}{0.334} & \scalebox{1.35}{0.438} & \scalebox{1.35}{0.369} & \scalebox{1.35}{0.473} \\
 & \scalebox{1.35}{60} & \scalebox{1.35}{\textbf{0.328}} & \scalebox{1.35}{0.412} & \scalebox{1.35}{0.342} & \scalebox{1.35}{0.422} & \scalebox{1.35}{0.348} & \scalebox{1.35}{0.432} & \scalebox{1.35}{0.362} & \scalebox{1.35}{0.441} & \scalebox{1.35}{0.389} & \scalebox{1.35}{0.473} & \scalebox{1.35}{0.360} & \scalebox{1.35}{0.446} & \scalebox{1.35}{\textbf{0.328}} & \scalebox{1.35}{0.440} & \scalebox{1.35}{0.366} & \scalebox{1.35}{0.471} \\
 & \scalebox{1.35}{90} & \scalebox{1.35}{\textbf{0.329}} & \scalebox{1.35}{0.418} & \scalebox{1.35}{0.378} & \scalebox{1.35}{0.454} & \scalebox{1.35}{0.368} & \scalebox{1.35}{0.445} & \scalebox{1.35}{0.374} & \scalebox{1.35}{0.449} & \scalebox{1.35}{0.422} & \scalebox{1.35}{0.491} & \scalebox{1.35}{0.380} & \scalebox{1.35}{0.461} & \scalebox{1.35}{0.331} & \scalebox{1.35}{0.452} & \scalebox{1.35}{0.371} & \scalebox{1.35}{0.476} \\
 & \scalebox{1.35}{120} & \scalebox{1.35}{0.388} & \scalebox{1.35}{0.469} & \scalebox{1.35}{0.414} & \scalebox{1.35}{0.487} & \scalebox{1.35}{0.438} & \scalebox{1.35}{0.499} & \scalebox{1.35}{0.406} & \scalebox{1.35}{0.475} & \scalebox{1.35}{0.458} & \scalebox{1.35}{0.517} & \scalebox{1.35}{0.396} & \scalebox{1.35}{0.472} & \scalebox{1.35}{\textbf{0.338}} & \scalebox{1.35}{0.458} & \scalebox{1.35}{0.377} & \scalebox{1.35}{0.482} \\
\cmidrule(lr){2-18}
 & \scalebox{1.35}{AVG} & \scalebox{1.35}{\textbf{0.286}} & \scalebox{1.35}{0.382} & \scalebox{1.35}{0.324} & \scalebox{1.35}{0.406} & \scalebox{1.35}{0.322} & \scalebox{1.35}{0.413} & \scalebox{1.35}{0.317} & \scalebox{1.35}{0.409} & \scalebox{1.35}{0.340} & \scalebox{1.35}{0.430} & \scalebox{1.35}{0.321} & \scalebox{1.35}{0.411} & \scalebox{1.35}{0.307} & \scalebox{1.35}{0.421} & \scalebox{1.35}{0.446} & \scalebox{1.35}{0.528} \\
\bottomrule[1.2pt]
\multirow{8}{*}{\scalebox{1.35}{\rotatebox{90}{ES-LJU}}}
 & \scalebox{1.35}{1} & \scalebox{1.35}{0.109} & \scalebox{1.35}{0.234} & \scalebox{1.35}{0.127} & \scalebox{1.35}{0.257} & \scalebox{1.35}{0.120} & \scalebox{1.35}{0.244} & \scalebox{1.35}{0.123} & \scalebox{1.35}{0.262} & \scalebox{1.35}{\textbf{0.099}} & \scalebox{1.35}{0.229} & \scalebox{1.35}{0.104} & \scalebox{1.35}{0.228} & \scalebox{1.35}{0.144} & \scalebox{1.35}{0.297} & \scalebox{1.35}{0.460} & \scalebox{1.35}{0.508} \\
 & \scalebox{1.35}{7} & \scalebox{1.35}{\textbf{0.147}} & \scalebox{1.35}{0.277} & \scalebox{1.35}{0.149} & \scalebox{1.35}{0.284} & \scalebox{1.35}{0.159} & \scalebox{1.35}{0.292} & \scalebox{1.35}{0.157} & \scalebox{1.35}{0.296} & \scalebox{1.35}{0.163} & \scalebox{1.35}{0.300} & \scalebox{1.35}{0.154} & \scalebox{1.35}{0.287} & \scalebox{1.35}{0.169} & \scalebox{1.35}{0.304} & \scalebox{1.35}{0.359} & \scalebox{1.35}{0.455} \\
 & \scalebox{1.35}{14} & \scalebox{1.35}{0.172} & \scalebox{1.35}{0.301} & \scalebox{1.35}{0.163} & \scalebox{1.35}{0.292} & \scalebox{1.35}{\textbf{0.171}} & \scalebox{1.35}{0.299} & \scalebox{1.35}{0.174} & \scalebox{1.35}{0.308} & \scalebox{1.35}{0.182} & \scalebox{1.35}{0.319} & \scalebox{1.35}{0.187} & \scalebox{1.35}{0.315} & \scalebox{1.35}{0.172} & \scalebox{1.35}{0.312} & \scalebox{1.35}{0.345} & \scalebox{1.35}{0.462} \\
 & \scalebox{1.35}{30} & \scalebox{1.35}{\textbf{0.187}} & \scalebox{1.35}{0.315} & \scalebox{1.35}{0.205} & \scalebox{1.35}{0.321} & \scalebox{1.35}{0.211} & \scalebox{1.35}{0.338} & \scalebox{1.35}{0.209} & \scalebox{1.35}{0.336} & \scalebox{1.35}{0.213} & \scalebox{1.35}{0.344} & \scalebox{1.35}{0.230} & \scalebox{1.35}{0.351} & \scalebox{1.35}{0.197} & \scalebox{1.35}{0.342} & \scalebox{1.35}{0.360} & \scalebox{1.35}{0.473} \\
 & \scalebox{1.35}{60} & \scalebox{1.35}{\textbf{0.203}} & \scalebox{1.35}{0.333} & \scalebox{1.35}{0.269} & \scalebox{1.35}{0.364} & \scalebox{1.35}{0.260} & \scalebox{1.35}{0.388} & \scalebox{1.35}{0.256} & \scalebox{1.35}{0.378} & \scalebox{1.35}{0.259} & \scalebox{1.35}{0.376} & \scalebox{1.35}{0.292} & \scalebox{1.35}{0.394} & \scalebox{1.35}{0.251} & \scalebox{1.35}{0.385} & \scalebox{1.35}{0.348} & \scalebox{1.35}{0.467} \\
 & \scalebox{1.35}{90} & \scalebox{1.35}{\textbf{0.245}} & \scalebox{1.35}{0.364} & \scalebox{1.35}{0.308} & \scalebox{1.35}{0.382} & \scalebox{1.35}{0.285} & \scalebox{1.35}{0.393} & \scalebox{1.35}{0.297} & \scalebox{1.35}{0.407} & \scalebox{1.35}{0.288} & \scalebox{1.35}{0.394} & \scalebox{1.35}{0.330} & \scalebox{1.35}{0.414} & \scalebox{1.35}{0.286} & \scalebox{1.35}{0.416} & \scalebox{1.35}{0.367} & \scalebox{1.35}{0.484} \\
 & \scalebox{1.35}{120} & \scalebox{1.35}{\textbf{0.270}} & \scalebox{1.35}{0.382} & \scalebox{1.35}{0.327} & \scalebox{1.35}{0.390} & \scalebox{1.35}{0.308} & \scalebox{1.35}{0.412} & \scalebox{1.35}{0.305} & \scalebox{1.35}{0.401} & \scalebox{1.35}{0.306} & \scalebox{1.35}{0.407} & \scalebox{1.35}{0.358} & \scalebox{1.35}{0.430} & \scalebox{1.35}{0.293} & \scalebox{1.35}{0.421} & \scalebox{1.35}{0.385} & \scalebox{1.35}{0.492} \\
\cmidrule(lr){2-18}
 & \scalebox{1.35}{AVG} & \scalebox{1.35}{\textbf{0.190}} & \scalebox{1.35}{0.315} & \scalebox{1.35}{0.221} & \scalebox{1.35}{0.327} & \scalebox{1.35}{0.216} & \scalebox{1.35}{0.338} & \scalebox{1.35}{0.217} & \scalebox{1.35}{0.341} & \scalebox{1.35}{0.216} & \scalebox{1.35}{0.338} & \scalebox{1.35}{0.236} & \scalebox{1.35}{0.346} & \scalebox{1.35}{0.216} & \scalebox{1.35}{0.354} & \scalebox{1.35}{0.375} & \scalebox{1.35}{0.477} \\
\bottomrule[1.2pt]

\end{tabular}}
\label{tab:flux_tower_results}
\end{sc}
\end{table*}

\begin{table*}[htbp]
\caption{Full results of the long-term prediction task for benchmark dataset with exogenous drivers. The lowest MSE for each horizon and dataset is \textbf{emphasized} in bold.}
\vspace{2pt}
\setlength{\tabcolsep}{2.5pt}
\renewcommand{\arraystretch}{1.5}
\begin{sc}
\resizebox{\linewidth}{!}{
\begin{tabular}{c|c|cc|cc|cc|cc|cc|cc|cc|cc|cc}
\toprule[1.2pt]  
\multicolumn{2}{c}{\multirow{2}{*}{\scalebox{1.35}{Models}}}   & \multicolumn{2}{c}{\scalebox{1.35}{ExoTST}} & \multicolumn{2}{c}{\scalebox{1.35}{iTrans.}} & \multicolumn{2}{c}{\scalebox{1.35}{(U)PatchTST}} & \multicolumn{2}{c}{\scalebox{1.35}{(M)PatchTST}} & \multicolumn{2}{c}{\scalebox{1.35}{DLinear} }& \multicolumn{2}{c}{\scalebox{1.35}{TiDE}} & \multicolumn{2}{c}{\scalebox{1.35}{ED-LSTM}} & \multicolumn{2}{c}{\scalebox{1.35}{LSTM}} \\ 
\multicolumn{2}{c}{} & \multicolumn{2}{c}{\scalebox{1.35}{(Ours)}} & \multicolumn{2}{c}{\scalebox{1.35}{\cite{liu2023itransformer}}} & \multicolumn{2}{c}{\scalebox{1.35}{\cite{nie2022time}}} & \multicolumn{2}{c}{\scalebox{1.35}{\cite{nie2022time}}} & \multicolumn{2}{c}{\scalebox{1.35}{\cite{zeng2023transformers}}} & \multicolumn{2}{c}{\scalebox{1.35}{\cite{das2023long}}} & \multicolumn{2}{c}{\scalebox{1.35}{\cite{yin2022rr}}} & \multicolumn{2}{c}{\scalebox{1.35}{\cite{kratzert2018rainfall}}}\\ 
\cmidrule(lr){3-4}\cmidrule(lr){5-6}\cmidrule(lr){7-8}\cmidrule(lr){9-10}\cmidrule(lr){11-12}\cmidrule(lr){13-14}\cmidrule(lr){15-16}\cmidrule(lr){17-18}
\multicolumn{2}{c}{\scalebox{1.35}{Metric}}   & \scalebox{1.35}{MSE} & \scalebox{1.35}{MAE} & \scalebox{1.35}{MSE}    & \scalebox{1.35}{MAE}   & \scalebox{1.35}{MSE}    & \scalebox{1.35}{MAE}    & \scalebox{1.35}{MSE}    & \scalebox{1.35}{MAE} & \scalebox{1.35}{MSE}    & \scalebox{1.35}{MAE} & \scalebox{1.35}{MSE}    & \scalebox{1.35}{MAE}   & \scalebox{1.35}{MSE}    & \scalebox{1.35}{MAE}  & \scalebox{1.35}{MSE}    & \scalebox{1.35}{MAE}      \\ 
\toprule[1.2pt] 
\multirow{8}{*}{\scalebox{1.35}{\rotatebox{90}{ETTh1}}}
 & \scalebox{1.35}{1} & \scalebox{1.35}{\textbf{0.006}} & \scalebox{1.35}{0.052} & \scalebox{1.35}{0.007} & \scalebox{1.35}{0.058} & \scalebox{1.35}{\textbf{0.006}} & \scalebox{1.35}{0.056} & \scalebox{1.35}{0.008} & \scalebox{1.35}{0.065} & \scalebox{1.35}{0.007} & \scalebox{1.35}{0.057} & \scalebox{1.35}{0.007} & \scalebox{1.35}{0.057} & \scalebox{1.35}{0.010} & \scalebox{1.35}{0.070} & \scalebox{1.35}{2.035} & \scalebox{1.35}{1.277} \\
 & \scalebox{1.35}{7} & \scalebox{1.35}{\textbf{0.020}} & \scalebox{1.35}{0.101} & \scalebox{1.35}{0.024} & \scalebox{1.35}{0.111} & \scalebox{1.35}{0.021} & \scalebox{1.35}{0.105} & \scalebox{1.35}{\textbf{0.020}} & \scalebox{1.35}{0.101} & \scalebox{1.35}{0.024} & \scalebox{1.35}{0.109} & \scalebox{1.35}{0.021} & \scalebox{1.35}{0.102} & \scalebox{1.35}{0.030} & \scalebox{1.35}{0.130} & \scalebox{1.35}{2.031} & \scalebox{1.35}{1.281} \\
 & \scalebox{1.35}{14} & \scalebox{1.35}{\textbf{0.032}} & \scalebox{1.35}{0.130} & \scalebox{1.35}{0.039} & \scalebox{1.35}{0.143} & \scalebox{1.35}{0.036} & \scalebox{1.35}{0.139} & \scalebox{1.35}{0.035} & \scalebox{1.35}{0.136} & \scalebox{1.35}{0.034} & \scalebox{1.35}{0.134} & \scalebox{1.35}{0.037} & \scalebox{1.35}{0.138} & \scalebox{1.35}{0.046} & \scalebox{1.35}{0.164} & \scalebox{1.35}{1.898} & \scalebox{1.35}{1.214} \\
 & \scalebox{1.35}{30} & \scalebox{1.35}{\textbf{0.044}} & \scalebox{1.35}{0.158} & \scalebox{1.35}{0.061} & \scalebox{1.35}{0.182} & \scalebox{1.35}{0.068} & \scalebox{1.35}{0.189} & \scalebox{1.35}{0.074} & \scalebox{1.35}{0.206} & \scalebox{1.35}{0.064} & \scalebox{1.35}{0.199} & \scalebox{1.35}{0.056} & \scalebox{1.35}{0.177} & \scalebox{1.35}{0.069} & \scalebox{1.35}{0.200} & \scalebox{1.35}{1.421} & \scalebox{1.35}{1.039} \\
 & \scalebox{1.35}{60} & \scalebox{1.35}{\textbf{0.067}} & \scalebox{1.35}{0.195} & \scalebox{1.35}{0.097} & \scalebox{1.35}{0.232} & \scalebox{1.35}{0.113} & \scalebox{1.35}{0.254} & \scalebox{1.35}{0.118} & \scalebox{1.35}{0.254} & \scalebox{1.35}{0.084} & \scalebox{1.35}{0.227} & \scalebox{1.35}{0.079} & \scalebox{1.35}{0.217} & \scalebox{1.35}{0.146} & \scalebox{1.35}{0.303} & \scalebox{1.35}{1.287} & \scalebox{1.35}{0.996} \\
 & \scalebox{1.35}{90} & \scalebox{1.35}{\textbf{0.086}} & \scalebox{1.35}{0.223} & \scalebox{1.35}{0.122} & \scalebox{1.35}{0.258} & \scalebox{1.35}{0.137} & \scalebox{1.35}{0.278} & \scalebox{1.35}{0.143} & \scalebox{1.35}{0.285} & \scalebox{1.35}{0.089} & \scalebox{1.35}{0.228} & \scalebox{1.35}{0.094} & \scalebox{1.35}{0.235} & \scalebox{1.35}{0.324} & \scalebox{1.35}{0.484} & \scalebox{1.35}{1.276} & \scalebox{1.35}{1.008} \\
 & \scalebox{1.35}{120} & \scalebox{1.35}{\textbf{0.087}} & \scalebox{1.35}{0.228} & \scalebox{1.35}{0.136} & \scalebox{1.35}{0.276} & \scalebox{1.35}{0.154} & \scalebox{1.35}{0.296} & \scalebox{1.35}{0.156} & \scalebox{1.35}{0.295} & \scalebox{1.35}{0.101} & \scalebox{1.35}{0.243} & \scalebox{1.35}{0.101} & \scalebox{1.35}{0.245} & \scalebox{1.35}{0.369} & \scalebox{1.35}{0.505} & \scalebox{1.35}{1.264} & \scalebox{1.35}{1.010} \\
\cmidrule(lr){2-18}
 & \scalebox{1.35}{AVG} & \scalebox{1.35}{\textbf{0.049}} & \scalebox{1.35}{0.155} & \scalebox{1.35}{0.069} & \scalebox{1.35}{0.180} & \scalebox{1.35}{0.076} & \scalebox{1.35}{0.188} & \scalebox{1.35}{0.079} & \scalebox{1.35}{0.192} & \scalebox{1.35}{0.058} & \scalebox{1.35}{0.171} & \scalebox{1.35}{0.056} & \scalebox{1.35}{0.167} & \scalebox{1.35}{0.142} & \scalebox{1.35}{0.265} & \scalebox{1.35}{1.602} & \scalebox{1.35}{1.118} \\
\bottomrule[1.2pt]
\multirow{8}{*}{\scalebox{1.35}{\rotatebox{90}{Exchange Rate}}}
 & \scalebox{1.35}{1} & \scalebox{1.35}{\textbf{0.004}} & \scalebox{1.35}{0.045} & \scalebox{1.35}{0.007} & \scalebox{1.35}{0.064} & \scalebox{1.35}{\textbf{0.004}} & \scalebox{1.35}{0.047} & \scalebox{1.35}{\textbf{0.004}} & \scalebox{1.35}{0.046} & \scalebox{1.35}{0.005} & \scalebox{1.35}{0.056} & \scalebox{1.35}{0.005} & \scalebox{1.35}{0.053} & \scalebox{1.35}{0.040} & \scalebox{1.35}{0.178} & \scalebox{1.35}{0.769} & \scalebox{1.35}{0.814} \\
 & \scalebox{1.35}{7} & \scalebox{1.35}{\textbf{0.011}} & \scalebox{1.35}{0.082} & \scalebox{1.35}{0.017} & \scalebox{1.35}{0.101} & \scalebox{1.35}{\textbf{0.011}} & \scalebox{1.35}{0.081} & \scalebox{1.35}{\textbf{0.011}} & \scalebox{1.35}{0.079} & \scalebox{1.35}{\textbf{0.011}} & \scalebox{1.35}{0.082} & \scalebox{1.35}{\textbf{0.011}} & \scalebox{1.35}{0.081} & \scalebox{1.35}{0.104} & \scalebox{1.35}{0.301} & \scalebox{1.35}{1.184} & \scalebox{1.35}{1.021} \\
 & \scalebox{1.35}{14} & \scalebox{1.35}{\textbf{0.017}} & \scalebox{1.35}{0.104} & \scalebox{1.35}{0.026} & \scalebox{1.35}{0.126} & \scalebox{1.35}{0.019} & \scalebox{1.35}{0.103} & \scalebox{1.35}{0.018} & \scalebox{1.35}{0.102} & \scalebox{1.35}{0.025} & \scalebox{1.35}{0.123} & \scalebox{1.35}{0.018} & \scalebox{1.35}{0.102} & \scalebox{1.35}{0.158} & \scalebox{1.35}{0.368} & \scalebox{1.35}{1.279} & \scalebox{1.35}{1.063} \\
 & \scalebox{1.35}{30} & \scalebox{1.35}{\textbf{0.028}} & \scalebox{1.35}{0.132} & \scalebox{1.35}{0.046} & \scalebox{1.35}{0.164} & \scalebox{1.35}{0.044} & \scalebox{1.35}{0.162} & \scalebox{1.35}{0.042} & \scalebox{1.35}{0.157} & \scalebox{1.35}{0.048} & \scalebox{1.35}{0.172} & \scalebox{1.35}{0.033} & \scalebox{1.35}{0.139} & \scalebox{1.35}{0.389} & \scalebox{1.35}{0.560} & \scalebox{1.35}{1.372} & \scalebox{1.35}{1.109} \\
 & \scalebox{1.35}{60} & \scalebox{1.35}{\textbf{0.056}} & \scalebox{1.35}{0.185} & \scalebox{1.35}{0.089} & \scalebox{1.35}{0.223} & \scalebox{1.35}{0.087} & \scalebox{1.35}{0.224} & \scalebox{1.35}{0.086} & \scalebox{1.35}{0.228} & \scalebox{1.35}{0.068} & \scalebox{1.35}{0.200} & \scalebox{1.35}{0.064} & \scalebox{1.35}{0.191} & \scalebox{1.35}{0.570} & \scalebox{1.35}{0.683} & \scalebox{1.35}{1.347} & \scalebox{1.35}{1.102} \\
 & \scalebox{1.35}{90} & \scalebox{1.35}{\textbf{0.068}} & \scalebox{1.35}{0.210} & \scalebox{1.35}{0.129} & \scalebox{1.35}{0.270} & \scalebox{1.35}{0.158} & \scalebox{1.35}{0.310} & \scalebox{1.35}{0.153} & \scalebox{1.35}{0.304} & \scalebox{1.35}{0.091} & \scalebox{1.35}{0.233} & \scalebox{1.35}{0.098} & \scalebox{1.35}{0.232} & \scalebox{1.35}{0.732} & \scalebox{1.35}{0.798} & \scalebox{1.35}{1.387} & \scalebox{1.35}{1.120} \\
 & \scalebox{1.35}{120} & \scalebox{1.35}{\textbf{0.095}} & \scalebox{1.35}{0.243} & \scalebox{1.35}{0.212} & \scalebox{1.35}{0.337} & \scalebox{1.35}{0.211} & \scalebox{1.35}{0.357} & \scalebox{1.35}{0.214} & \scalebox{1.35}{0.358} & \scalebox{1.35}{0.143} & \scalebox{1.35}{0.290} & \scalebox{1.35}{0.132} & \scalebox{1.35}{0.270} & \scalebox{1.35}{0.830} & \scalebox{1.35}{0.840} & \scalebox{1.35}{1.132} & \scalebox{1.35}{1.011} \\
\cmidrule(lr){2-18}
 & \scalebox{1.35}{AVG} & \scalebox{1.35}{\textbf{0.040}} & \scalebox{1.35}{0.143} & \scalebox{1.35}{0.075} & \scalebox{1.35}{0.184} & \scalebox{1.35}{0.076} & \scalebox{1.35}{0.183} & \scalebox{1.35}{0.075} & \scalebox{1.35}{0.182} & \scalebox{1.35}{0.056} & \scalebox{1.35}{0.165} & \scalebox{1.35}{0.052} & \scalebox{1.35}{0.153} & \scalebox{1.35}{0.403} & \scalebox{1.35}{0.533} & \scalebox{1.35}{1.210} & \scalebox{1.35}{1.034} \\
\bottomrule[1.2pt]
\multirow{8}{*}{\scalebox{1.35}{\rotatebox{90}{ECL}}}
 & \scalebox{1.35}{1} & \scalebox{1.35}{0.052} & \scalebox{1.35}{0.162} & \scalebox{1.35}{0.052} & \scalebox{1.35}{0.160} & \scalebox{1.35}{\textbf{0.048}} & \scalebox{1.35}{0.150} & \scalebox{1.35}{0.059} & \scalebox{1.35}{0.166} & \scalebox{1.35}{0.049} & \scalebox{1.35}{0.143} & \scalebox{1.35}{0.055} & \scalebox{1.35}{0.158} & \scalebox{1.35}{0.142} & \scalebox{1.35}{0.284} & \scalebox{1.35}{0.368} & \scalebox{1.35}{0.478} \\
 & \scalebox{1.35}{7} & \scalebox{1.35}{0.121} & \scalebox{1.35}{0.258} & \scalebox{1.35}{0.113} & \scalebox{1.35}{0.247} & \scalebox{1.35}{\textbf{0.108}} & \scalebox{1.35}{0.237} & \scalebox{1.35}{0.121} & \scalebox{1.35}{0.253} & \scalebox{1.35}{0.109} & \scalebox{1.35}{0.231} & \scalebox{1.35}{0.114} & \scalebox{1.35}{0.240} & \scalebox{1.35}{0.292} & \scalebox{1.35}{0.419} & \scalebox{1.35}{0.351} & \scalebox{1.35}{0.472} \\
 & \scalebox{1.35}{14} & \scalebox{1.35}{\textbf{0.130}} & \scalebox{1.35}{0.271} & \scalebox{1.35}{0.140} & \scalebox{1.35}{0.274} & \scalebox{1.35}{0.169} & \scalebox{1.35}{0.297} & \scalebox{1.35}{0.140} & \scalebox{1.35}{0.272} & \scalebox{1.35}{0.139} & \scalebox{1.35}{0.265} & \scalebox{1.35}{0.145} & \scalebox{1.35}{0.271} & \scalebox{1.35}{0.301} & \scalebox{1.35}{0.429} & \scalebox{1.35}{0.376} & \scalebox{1.35}{0.489} \\
 & \scalebox{1.35}{30} & \scalebox{1.35}{\textbf{0.146}} & \scalebox{1.35}{0.290} & \scalebox{1.35}{0.183} & \scalebox{1.35}{0.311} & \scalebox{1.35}{0.200} & \scalebox{1.35}{0.329} & \scalebox{1.35}{0.174} & \scalebox{1.35}{0.304} & \scalebox{1.35}{0.167} & \scalebox{1.35}{0.293} & \scalebox{1.35}{0.173} & \scalebox{1.35}{0.300} & \scalebox{1.35}{0.324} & \scalebox{1.35}{0.451} & \scalebox{1.35}{0.490} & \scalebox{1.35}{0.564} \\
 & \scalebox{1.35}{60} & \scalebox{1.35}{\textbf{0.169}} & \scalebox{1.35}{0.314} & \scalebox{1.35}{0.244} & \scalebox{1.35}{0.356} & \scalebox{1.35}{0.263} & \scalebox{1.35}{0.368} & \scalebox{1.35}{0.263} & \scalebox{1.35}{0.366} & \scalebox{1.35}{0.210} & \scalebox{1.35}{0.332} & \scalebox{1.35}{0.214} & \scalebox{1.35}{0.332} & \scalebox{1.35}{0.357} & \scalebox{1.35}{0.477} & \scalebox{1.35}{0.399} & \scalebox{1.35}{0.494} \\
 & \scalebox{1.35}{90} & \scalebox{1.35}{\textbf{0.179}} & \scalebox{1.35}{0.322} & \scalebox{1.35}{0.268} & \scalebox{1.35}{0.375} & \scalebox{1.35}{0.365} & \scalebox{1.35}{0.431} & \scalebox{1.35}{0.395} & \scalebox{1.35}{0.453} & \scalebox{1.35}{0.231} & \scalebox{1.35}{0.346} & \scalebox{1.35}{0.230} & \scalebox{1.35}{0.341} & \scalebox{1.35}{0.371} & \scalebox{1.35}{0.490} & \scalebox{1.35}{0.422} & \scalebox{1.35}{0.515} \\
 & \scalebox{1.35}{120} & \scalebox{1.35}{\textbf{0.180}} & \scalebox{1.35}{0.321} & \scalebox{1.35}{0.276} & \scalebox{1.35}{0.377} & \scalebox{1.35}{0.397} & \scalebox{1.35}{0.449} & \scalebox{1.35}{0.493} & \scalebox{1.35}{0.513} & \scalebox{1.35}{0.234} & \scalebox{1.35}{0.344} & \scalebox{1.35}{0.238} & \scalebox{1.35}{0.345} & \scalebox{1.35}{0.381} & \scalebox{1.35}{0.505} & \scalebox{1.35}{0.445} & \scalebox{1.35}{0.522} \\
\cmidrule(lr){2-18}
 & \scalebox{1.35}{AVG} & \scalebox{1.35}{\textbf{0.140}} & \scalebox{1.35}{0.277} & \scalebox{1.35}{0.182} & \scalebox{1.35}{0.300} & \scalebox{1.35}{0.221} & \scalebox{1.35}{0.323} & \scalebox{1.35}{0.235} & \scalebox{1.35}{0.332} & \scalebox{1.35}{0.163} & \scalebox{1.35}{0.279} & \scalebox{1.35}{0.167} & \scalebox{1.35}{0.284} & \scalebox{1.35}{0.310} & \scalebox{1.35}{0.436} & \scalebox{1.35}{0.407} & \scalebox{1.35}{0.505} \\
\bottomrule[1.2pt]
\end{tabular}}
\label{tab:benchmark_dataset}
\end{sc}
\end{table*}

\subsection{Carbon Flux Data}

\textbf{Data Description} 
In this study, we use Flux Tower Dataset ~\cite{pastorello2020fluxnet2015}, which is a global network of eddy-covariance stations that record carbon, water, and energy exchanges between the atmosphere and the biosphere. It offers valuable ecosystem-scale observations across various climate and ecosystem types. In this dataset, our goal is to predict Gross Primary Productivity (GPP) on a daily timescale. We achieve this by using a combination of exogenous meteorological and remote sensing inputs, including precipitation, 2-meter air temperature (Ta), vapor pressure deficit (VPD), incoming short-wave radiation from ERA5 reanalysis data~\cite{hersbach2020era5}, and Leaf Area Index (LAI) data derived from the MODIS remote sensing data~\cite{mkp15}, an approach successfully employed in previous studies ~\cite{renganathan2023task, nathaniel2023metaflux}. In our study, we focus on three distinct sites, each representing a unique eco-region: DE-HAI (deciduous forest, Germany), ES-LJU (open shrubland, Spain) and AT-NEU (managed grassland, Austria). Each site was selected for its distinctive ecological processes and interactions, which are characteristic of their respective eco-regions. This approach tests our methodology's robustness across various environmental conditions and its general applications. Figure \ref{fig:gpp_illustration} illustrates the daily Gross Primary Productivity (GPP) for ES-LJU. The plot highlights the complex temporal patterns in GPP data. As we can observe, despite some seasonality and underlying trends, predicting GPP using only past GPP (endogenous response) is insufficient. This underscores the importance of exogenous drivers that drive the GPP response, and hence, incorporating current drivers that influence GPP becomes essential for accurate GPP predictions.

\begin{figure}[t]
    \centering
    \includegraphics[width=0.9\linewidth]{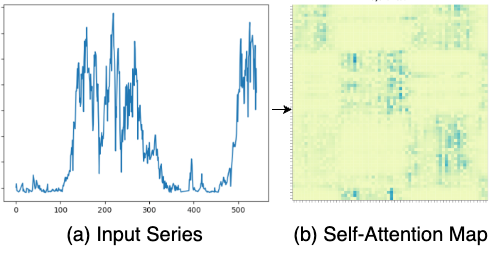} 
    
    \caption{\footnotesize Visualization of the learned self-attention map alongside the endogenous look-back window for interpretation.}
    \label{fig:self_attention}
    
\end{figure}

\textbf{Results}
We present Mean Squared Error (MSE) and Mean Absolute Error (MAE) for all datasets and methods in Table \ref{tab:flux_tower_results}. Bold-faced values indicate the best model in each row. Predictions are reported for various horizons (1, 7, 14, 30, 60, 90, and 120 days) to demonstrate model performance across both short-term and long-term forecasts.

A one-day prediction represents daily real-life scenarios where past endogenous and exogenous variables are combined with today's exogenous drivers. For example, predicting today's GPP based on historical GPP data, past weather conditions, and today's meteorological forecasts. Shorter horizons (1, 7, 14, and 30 days) test the models' ability to leverage past observations and current exogenous conditions for accurate short-term predictions used in many operational decision-making and real-time monitoring. Longer horizons (60, 90, and 120 days) evaluate the models' long-term prediction skills, which are essential for applications like GPP or flood prediction and require incorporating different future scenarios of meteorological conditions.

From the results in Table \ref{tab:flux_tower_results}, we have the following high-level observations: (a)
ExoTST consistently achieves the lowest or near lowest MSE \& MAE across all prediction horizons, showing its capability to integrate available exogenous information and improve prediction accuracy effectively. Specifically, ExoTST outperforms all baseline models in MSE by a margin of up to 10\%.
(b) LSTM, a forward model, significantly underperforms at all prediction horizons despite using current exogenous information due to its inability to capture long-term dependencies. This highlights the importance of effectively integrating past context.(c) The ED-LSTM and TiDE models, which utilize all available exogenous drivers and past context, demonstrate superior performance for short-term predictions. However, they tend to underperform at longer prediction horizons due to challenges in capturing long-term dependencies and extracting complex relationships between driver and response.
(d) Autoregressive models such as iTrans, (U)PatchTST, and (M)PatchTST are competitive at short intervals, showcasing the advantages of using attention mechanisms to learn temporal dependencies within the data effectively. This allows them to make accurate predictions in shorter horizons. However, their performance diminishes over longer horizons due to their inability to incorporate all exogenous drivers, which becomes crucial for longer horizons.

\textbf{Explainability}
Moreover, our model also offers interpretability due to its ability to visualize attention maps showing the temporal relationships for both exogenous and endogenous time series. Here (Figure \ref{fig:self_attention}), we visualize the attention map from the decoder. The self-attention map highlights four blue regions indicating different seasonal patterns within the series, suggesting that the model learns from similar seasonal fluctuations.  
These heatmaps demonstrate ExoTST's effectiveness in capturing temporal dependencies through its attention mechanisms and further validate that ExoTST is not merely reacting to random fluctuations but is understanding and leveraging repeating patterns for accurate predictions.

\subsection{Benchmark Data}
We also adopt three widely recognized public benchmarks for long-term multivariate forecasting to perform prediction with exogenous variables. We refer the reader to \cite{wu2021autoformer} for a detailed discussion of these datasets. Below, we detail the domain information for the endogenous drivers and exogenous responses for each scenario:

\textbf{Data Description}: \textit{Electricity Transformer Temperature (ETTh1)}: dataset contains two years of electric power data from a county in China, recorded hourly (H). Each data point includes an "oil temperature" measurement and six power load features categorized by usage frequency and intensity: High Usage Frequency Load (HUFL), High Usage Low Load (HULL), Medium Usage Frequency Load (MUFL), Medium Usage Low Load (MULL), Low Usage Frequency Load (LUFL), and Low Usage Low Load (LULL). Here, oil temperature is the endogenous response, while the six power load features act as exogenous drivers. The \textit{Electricity (ECL)} dataset comprises hourly electricity consumption records (in kWh) for 321 consumers over a three-year period. We predict the electricity consumption of the last consumer as the endogenous response, with the consumption data from the remaining 320 consumers serving as exogenous drivers. The \textit{Exchange dataset} includes daily exchange rates for eight foreign countries—Australia, Britain, Canada, Switzerland, China, Japan, New Zealand, and Singapore—from 1990 to 2010. The exchange rates of seven countries are treated as exogenous drivers, while the exchange rate of the eighth country serves as the endogenous response.

\textbf{Results}:
We present MSE and MAE for benchmark datasets in Table \ref{tab:benchmark_dataset}. As with GPP data, we report predictions for various horizons (1, 7, 14, 30, 60, 90, and 120 days) to demonstrate our model performance across both short-term and long-term forecasts. The results in Table \ref{tab:benchmark_dataset}
yield similar high-level observations as those from the Carbon Flux GPP Experiment, albeit with some notable differences. One key distinction is the significant challenge LSTM faces in learning forward models between exogenous drivers and responses, reflecting the datasets' heavy dependence on past context. TiDE outperforms all transformer-based approaches except our approach due to its ability to access all exogenous drivers.  While ED-LSTM also utilizes all exogenous drivers, it markedly underperforms compared to all the other non-LSTM baseline approaches due to LSTM's inability to retain long-term dependencies within the context effectively. 
DLinear consistently outperforms other transformer-based methods, as also evidenced in \cite{zeng2023transformers}. However, Despite Dlinear outperforming all the other transformer-based approaches, our method outperforms Dlinear due to effectively retaining complex temporal dependencies and learning the relationship between drivers and response.



\subsection{Robustness to missing value and noise}
In real-world scenarios, exogenous drivers originate from various sources and may have different frequencies. Synchronizing them is a challenge, and even after synchronization, they may be misaligned. Our model, by design, can handle misaligned/missing values from the exogenous data, and to simulate such misalignment, we intentionally introduce missing values in exogenous drivers. 


Table \ref{tab:missing} assesses models' robustness against missing values on the DE-HAI dataset. The results show the prediction error (average MSE) when 40\% and 80\% of the exogenous driver values are randomly masked (treated as missing) to simulate misalignment in these time series. Additionally, it analyzes performance differences between initial and later predictions by separately reporting the first and last 50 time steps, accounting for the impact of past context on predictions closer in time.

Key observations from the results indicate that ExoTST consistently achieves the lowest prediction error compared to models like TiDE, ED-LSTM, and LSTM across both percentages of missing values. As the percentage of missing exogenous values increases, the prediction error increases for all models. However, ExoTST remains the most robust, exhibiting the smallest increase in error.
TiDE, an MLP-based approach designed to handle exogenous drivers along with the past context, performs better than LSTM and ED-LSTM but is outperformed by ExoTST, especially as the missing values increase. ExoTST also maintains low errors across the first 50 and last 50 time steps. In contrast, LSTM, which relies heavily on exogenous drivers, shows a significant increase in error when more values are missing, highlighting its sensitivity to missing data.
The performance gap between TiDE and ExoTST is further highlighted when examining errors for the first 50 and last 50 time steps separately. With an 80\% missing fraction, ExoTST maintains low errors of 0.132 and 0.121 for the initial and later periods, respectively. In contrast, TiDE's error deteriorates from 0.169 to 0.189 between the two periods, indicating difficulties in leveraging past context effectively when future exogenous information is lacking. 
\begin{table}[]
\centering
\caption{\small Results for missing values in exogenous driver for DE-HAI}
\resizebox{0.45\textwidth}{!}{
\begin{tabular}{ccccc}
\toprule
\textbf{Model} & \boldmath{$\%$} & \multicolumn{3}{c}{\textbf{Prediction Error (Avg)}} \\
\cmidrule(lr){3-5}
& & \textbf{All Value} & \textbf{First 50}  & \textbf{Last 50} \\  
\midrule
ExoTST & 0.4 & 0.124 & 0.116 & 0.132 \\
TiDE  & 0.4 & 0.173 & 0.167 & 0.176 \\
ED-LSTM  & 0.4 & 0.180 & 0.208 & 0.159 \\
LSTM & 0.4 & 0.189 & 0.222 & 0.166 \\
\midrule
ExoTST & 0.8 & 0.127 & 0.132 & 0.121 \\
TiDE  & 0.8 & 0.179 & 0.169 & 0.189 \\
ED-LSTM  & 0.8 & 0.223 & 0.273 & 0.187 \\
LSTM & 0.8 & 0.217 & 0.265 & 0.183 \\

\bottomrule 
\end{tabular}}
\label{tab:missing}
\vspace{0.2cm}
\end{table}

\begin{table}[]
\centering
\caption{\small Results for noise in exogenous driver for DE-HAI}
\resizebox{0.45\textwidth}{!}{
\begin{tabular}{ccccc}
\toprule
\textbf{Model} & \boldmath{$\sigma$} & \multicolumn{3}{c}{\textbf{Prediction Error (Avg)}} \\
\cmidrule(lr){3-5}
& & \textbf{All Value} & \textbf{First 50}  & \textbf{Last 50} \\  
\midrule
ExoTST & 0.8 & 0.130 & 0.129 & 0.131 \\
TiDE & 0.8 & 0.175 & 0.169 & 0.179 \\
ED-LSTM & 0.8 & 0.189 & 0.211 & 0.174 \\
LSTM & 0.8 & 0.205 & 0.220 & 0.193 \\
\midrule
ExoTST & 1.2 & 0.139 & 0.139 & 0.139 \\
TiDE & 1.2 & 0.184 & 0.176 & 0.189 \\
ED-LSTM  & 1.2 & 0.241 & 0.244 & 0.242 \\
LSTM & 1.2 & 0.252 & 0.272 & 0.240 \\

\bottomrule 
\end{tabular}}
\label{tab:noise}
\vspace{0.2cm}
\end{table}


Moreover, ensuring continuous access to high-quality, noise-free exogenous data is challenging, particularly for real-time forecasting, given measurement errors, sensor limitations, and environmental conditions. To simulate such challenges, we assess ExoTST's robustness against noise in exogenous drivers by introducing Additive Gaussian Noise with zero mean and varying standard deviations($\sigma$) into the exogenous variables.

The results in Table \ref{tab:noise} show the robustness of ExoTST and other top baseline models against noise in the exogenous driver data for the DE-HAI dataset. Similar to the earlier results for missing value in Table \ref{tab:missing}, ExoTST consistently achieves the lowest prediction error compared to TiDE, ED-LSTM, and LSTM when Additive Gaussian Noise with varying standard deviations ($\sigma$ = 0.8 and 1.2) is introduced. In the presence of noise, ED-LSTM outperforms LSTM, suggesting its ability to leverage past context makes it more robust to noisy exogenous inputs. ExoTST maintains consistently low errors across both the first 50 and last 50 time steps, even at higher noise levels, while the deterioration in errors between the initial and later periods is more pronounced for TiDE, ED-LSTM, and LSTM when exogenous information is corrupted by noise. 

Overall, ExoTST's robust performance is attributed to its transformer-based architecture, which captures long-range dependencies and integrates context with available exogenous drivers through Cross-Temporal Modality Fusion and attention mechanisms, while TiDE's and ED-LSTM's concatenation-based methods struggle when exogenous data are noisy or partially missing.

\section{Conclusions}
In this paper, we highlight the importance of incorporating current/projected exogenous information for long-term time series prediction. To incorporate this information, we propose ExoTST, a novel transformer-based framework that effectively integrates current exogenous variables with historical context. ExoTST treats past and current/projected exogenous series as distinct modalities and introduces a cross-temporal modality fusion module to capture complex dependencies and interactions between drivers across time. We validated ExoTST effectiveness on both real-world carbon flux datasets and time series benchmarks, demonstrating significant improvement over the baseline. Moreover, our model exhibits the capacity to serve as a building block for a foundational model for time series prediction with current/projected exogenous drivers 



We note that the proposed method, ExoTST, is general and can add value across various domains where exogenous variables play a crucial role in prediction. For example, in the energy sector, predicting electricity loads is sensitive to varying weather conditions, such as temperature and humidity. By incorporating these exogenous factors, ExoTST can provide more accurate predictions, enabling better resource planning and allocation. Similarly, in retail and supply chain management, ExoTST can integrate market dynamics, economic cycles, and external events such as holidays and promotions with historical sales data to enhance demand forecasting, thereby optimizing inventory management and reducing costs.

\section*{Acknowledgment}


DL's and KT's work was supported by Dan Lu's Early Career Project, sponsered by the U.S. DOE's Office of Biological and Environmental Research. AR and VK were supported by the NSF LEAP Science and Technology Center (award 2019625) and the NSF grant (2313174). Most research was conducted at ORNL, operated by UT Battelle under DOE Contract DE-AC05-00OR22725, with computational resources provided by the Minnesota Supercomputing Institute.

\bibliographystyle{plain}
\bibliography{ref}

\end{document}